\documentclass[letterpaper, 10 pt, conference]{IEEEtran}  
\usepackage[utf8]{inputenc}

\usepackage{times}
\usepackage{graphicx}
\usepackage{epstopdf}

\usepackage{mdwmath}
\usepackage{mdwtab}
\usepackage{rotating}
\usepackage{caption}
\usepackage{subcaption}

\usepackage{amssymb}
\usepackage{amsfonts}
\usepackage{amsmath}
\usepackage{amsthm}
\usepackage{bm}

\usepackage{algorithm}
\usepackage{cite}
\usepackage{algorithmic}

\usepackage{multirow} 
\usepackage{multicol}
\usepackage{array}

\usepackage{standalone}
\usepackage{booktabs} 

\usepackage{siunitx} 

\usepackage{accents}

\usepackage{pgfplotstable} 
\usepackage{verbatim}
\usepackage{isomath}
\usepackage{overpic}

\usepackage{tikz}
\usetikzlibrary{shapes,calc,patterns,
	decorations.pathmorphing,
	decorations.markings}

\tikzstyle{spring}=[very thick,decorate,decoration={zigzag,pre length=2,post
	length=2,segment length=6}]

\tikzstyle{damper}=[thick,decoration={markings, 
	mark connection node=dmp,
	mark=at position 0.5 with 
	{
		\node (dmp) [thick,inner sep=0pt,transform shape,rotate=-90,minimum
		width=15pt,minimum height=3pt,draw=none] {};
		\draw [thick] ( $(dmp.north east)+(2pt,0)$ ) -- (dmp.south east) -- (dmp.south
		west) -- ( $(dmp.north west)+(2pt,0)$ );
		\draw [thick] ( $(dmp.north)+(0,-5pt)$ ) -- ( $(dmp.north)+(0,5pt)$ );
	}
}, decorate]

\makeatletter\newcommand{\manuallabel}[2]{\def\@currentlabel{#2}\label{#1}}\makeatother

\usepackage{color}
\usepackage{xcolor}

\colorlet   {lightorange}{orange!20}
\colorlet   {lightgrey}  {gray!20}

\usepackage{amssymb}
\usepackage{mathtools}

\mathchardef\mhyphen="2D   

\newcommand{\RNum}[1]{\uppercase\expandafter{\romannumeral #1\relax}}

\graphicspath{
{figures/}
}

\providecommand{\figurename}{Fig.}

\usepackage[inline]{enumitem}
\setlist{nolistsep}

\usepackage[normalem]{ulem}                                        
\usepackage{marginnote}
\usepackage[textwidth=10ex,colorinlistoftodos]{todonotes}

\colorlet{fwu}{red}
\colorlet{ywu}{blue}
\colorlet{zbing}{green}

\usepackage{graphics} 
\usepackage{graphicx}
\usepackage{epsfig} 
\usepackage{times} 
\usepackage{physics}
\usepackage{xcolor}
\usepackage{tikz}

\usetikzlibrary{calc} 
\usetikzlibrary{positioning}
\usepackage{balance}
\usepackage{soul}
\usetikzlibrary{arrows}
\usepackage{amsmath}
\usepackage{diagbox}
\usepackage[hidelinks]{hyperref}
\hypersetup{
    colorlinks=true,
    linkcolor=black,
    urlcolor=blue,
    citecolor=black
}

\usepackage[top=2.54cm, bottom=1.91cm, left=1.91cm, right=1.91cm]{geometry}

\IEEEoverridecommandlockouts

\title{\LARGE \bf%
VT-Bridge: Bridging Pretrained Foundation VLAs to VTLAs via Lightweight Residual Adaptation
}

\author{
Yansong Wu$^{1}$, Tuo Yang$^{1}$, Rongping Zhao$^{1}$, Lingyun Chen$^{1,2}$, Xiao Chen$^{1}$, \\Junnan Li$^{1}$, Fan Wu$^{3}$, Alois Knoll$^{1}$
\thanks{$^{1}$ Technical University of Munich, Germany. 
$^{2}$ Mohamed Bin Zayed University of Artificial Intelligence, Abu Dhabi, UAE.
$^{3}$ Shanghai University, China.
}%
}

\renewcommand{\baselinestretch}{0.96}

\let\oldtwocolumn\twocolumn
\renewcommand\twocolumn[1][]{%
    \oldtwocolumn[{#1}{
    \vspace{-15pt}
    \begin{center}
           \includegraphics[width=0.85\textwidth]{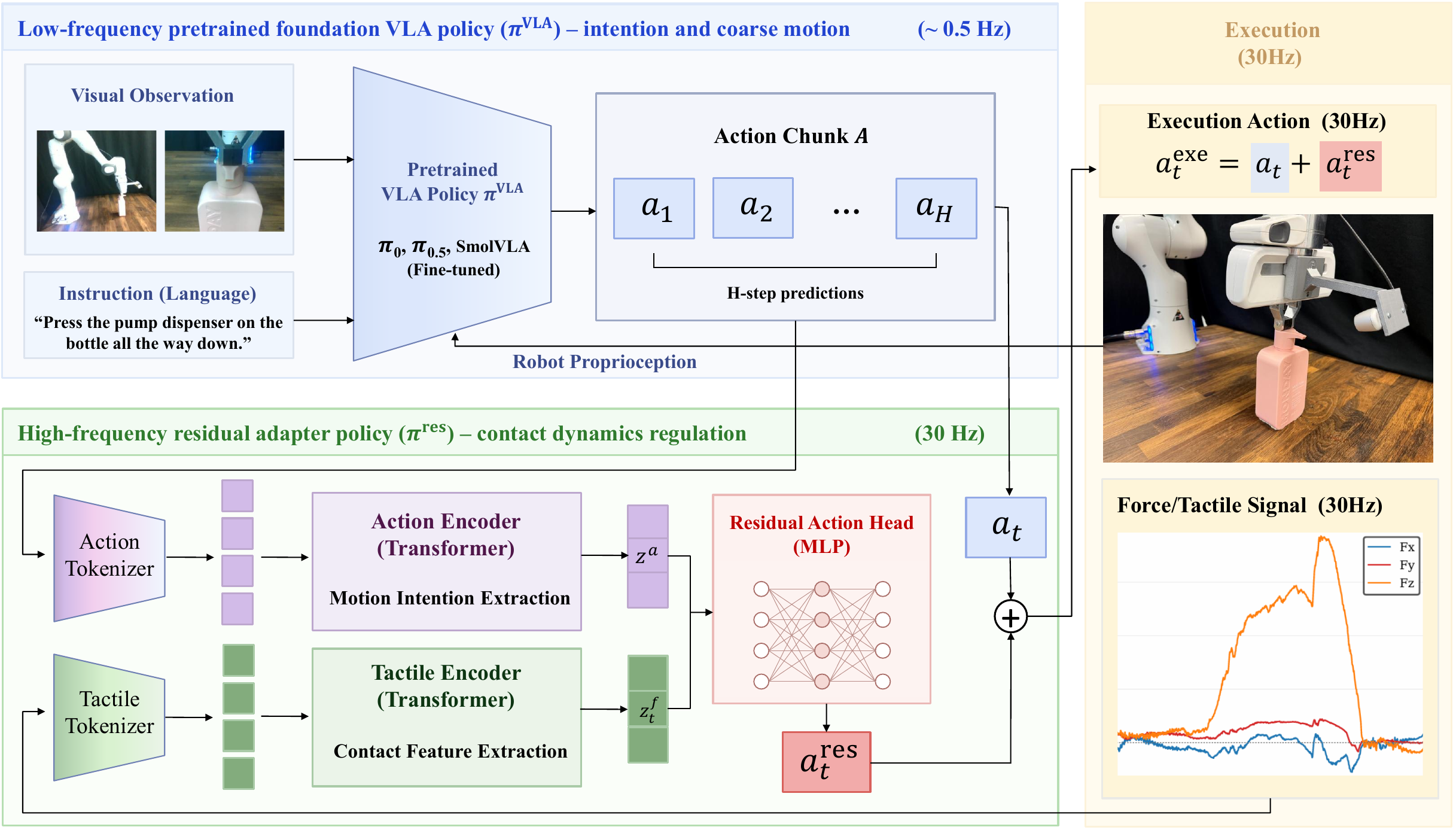}
           \captionof{figure}{Overview of VT-Bridge, a resource-efficient approach for bridging pretrained VLAs to VTLAs. It refines VLA-predicted actions using tactile feedback at the robot execution frequency. The residual-adapter architecture is applicable to different VLA backbones, with only its weights trained separately for each backbone.}
           \label{fig:system_overview}
    \end{center}
    \vspace{5pt}
    }]
}

\begin{document}
\maketitle

\begin{abstract}
Vision-Tactile-Language-Action (VTLA) models have demonstrated clear advantages over Vision-Language-Action (VLA) models in contact-rich manipulation. However, developing VTLA models is severely constrained by the massive amounts of vision-tactile data and computational resources required. To address this bottleneck, we propose VT-Bridge, a lightweight residual adaptation strategy that bridges pretrained foundation VLAs to VTLAs. Rather than training a VTLA model from scratch or modifying the original architecture of a pretrained VLA, VT-Bridge employs an identical lightweight residual-adapter architecture across VLA backbones and uses backbone-specific weights to refine actions at the robot execution frequency. This design substantially lowers the data and training barriers. Specifically, it requires up to 50 vision-tactile demonstrations per task to fine-tune a VLA backbone and train a 0.98M-parameter residual adapter. Experiments with three representative VLA backbones ($\pi_0$, $\pi_{0.5}$, and SmolVLA) across four contact-rich manipulation tasks further demonstrate its consistent effectiveness across VLA architectures. On average, VT-Bridge raises the task completion rate from 11.7\% with task-level VLA fine-tuning alone to 62.9\%. Together, these findings demonstrate the broad applicability, effectiveness, and accessibility of VT-Bridge for contact-rich manipulation. The project page is available at this \href{https://hoxnocha.github.io/vt-bridge-web/}{URL}.


\end{abstract}

\section{Introduction}
Vision-Language-Action (VLA) models illuminate a viable pathway towards general-purpose robot intelligence~\cite{firoozi2025foundation}, where a single pretrained policy can interpret natural-language instructions and execute diverse tasks across manipulation scenarios. Leveraging large-scale pretraining, VLA models show promising capabilities in tackling various tasks either zero-shot or through fine-tuning on a limited number of demonstrations~\cite{black2024pi0,black2025pi05,brohan2022rt1,chi2023diffusionpolicy,driess2023palme,guo2024prediction,kim2024openvla,kim2025finetuning,liu2025hybridvla,nvidia2025gr00tn1,octo2024,pertsch2025fast,zhou2025chatvla,zitkovich2023rt2}. Yet, even after task-specific fine-tuning, current VLAs continue to exhibit low success rates in real-world contact-rich manipulation, indicating that relying solely on non-contact modalities (vision, language, and robot kinematic  state) for action prediction is insufficient for reliable physical interaction.

To address this limitation, recent studies have begun incorporating tactile feedback into VLA architectures, giving rise to Vision-Tactile-Language-Action (VTLA) models~\cite{niu2026t,li2026favla,zhang2026craft,zhao2026fdvla,zhang2026feeling,li2026atvla,li2026forcevla2,yu2025forcevla,wang2026never,zhang2026unitacvla,bi2025vla,zhang2025vtla,zhang2026compliantvla,gubernatorov2026hapticvla,cheng2025omnivtla,zhang2025tavla,wang2026tacmamba,zhang2026tacvla,huang2025tactilevla}. These efforts span varied fusion strategies and model architectures, yielding promising gains in contact-rich manipulation tasks. Moreover, some empirical ablation studies have isolated tactile contributions, underscoring the critical role of tactile sensing in reliable physical interaction~\cite{li2026atvla,li2026forcevla2,yu2025forcevla}.


However, developing VTLA models remains costly, with substantial resource demands in both data collection and model training.
From the data perspective, paired vision-tactile trajectories are far scarcer than vision-only data in public robot datasets~\cite{fang2024rh20t,li2026forcevla2}. Moreover, collecting high-quality tactile demonstrations requires fine force regulation, significantly elevating teleoperation complexity~\cite{sliwowski2025reassemble, wu2025sharedassembly}. From the model perspective, tactile encoders and cross-modal fusion modules
introduce additional parameters and make the architecture more complex. These additions can also increase training costs. Constrained by these data and model bottlenecks, existing VTLAs commonly cover a significantly narrower range of tasks than general-purpose VLAs like $\pi_0$~\cite{black2024pi0}. This underscores a central question: \textit{Can we adapt existing open-source VLAs into VTLAs at a manageable cost?}

To address this question, we propose VT-Bridge, a lightweight residual adaptation strategy that bridges pretrained foundation VLAs to tactile-aware VTLAs for contact-rich manipulation. As shown in Fig.~\ref{fig:system_overview}, VT-Bridge introduces a lightweight residual adapter without modifying the pretrained VLA architecture. Based on the latest tactile feedback, the adapter refines each VLA-predicted action immediately before execution, thereby regulating contact dynamics at run-time.
This design retains the general knowledge acquired through large-scale VLA pretraining, while enabling the tactile responsiveness required for contact-rich manipulation. Moreover, VT-Bridge supports various pretrained VLA backbones using an identical residual adapter architecture, with the adapter weights tailored to each backbone. It significantly reduces the resources required to build VTLA policies for downstream contact-rich tasks. 




VT-Bridge possesses several key advantages:
\begin{itemize}
    \item \textbf{Pretrained VLAs Reuse and Compatibility:} 
    Instead of training a monolithic VTLA model from scratch, VT-Bridge preserves the pretrained VLA model and augments a pretrained VLA with a lightweight, tactile-conditioned residual adapter.  This design retains the vision-language grounding and motion-generation capabilities acquired during large-scale VLA pretraining. The same adapter architecture can be applied to different VLA backbones by tailoring its parameters to each backbone, as demonstrated with $\pi_0$~\cite{black2024pi0}, $\pi_{0.5}$~\cite{black2025pi05}, and SmolVLA~\cite{shukor2025smolvla}.

    \item \textbf{Low Data and Training Barriers:} Compared with training a VTLA model from scratch, VT-Bridge substantially lowers both data and training barriers. Specifically, it requires up to 50 vision-tactile demonstrations per task to fine-tune a VLA backbone and train a lightweight residual adapter with only 0.98M parameters. No additional large-scale training resources beyond those required for VLA fine-tuning are needed.
    \item \textbf{Run-time Reactivity:} As illustrated in Fig.~\ref{fig:system_overview}, VT-Bridge adopts a slow--fast inference scheme that performs semantic task reasoning and tactile adaptation at different frequencies. The VLA operates at a lower frequency to interpret complex scenes and generate action chunks encoding task intent and coarse motion, while the lightweight residual adapter responds to contact changes at the execution rate. 
    \item \textbf{Substantial and Consistent Performance Gains:} Despite its lightweight design and limited data requirements, VT-Bridge raises the average task completion rate from 11.7\% under task-level VLA fine-tuning to 62.9\%. The improvements are consistently observed across four real-world contact-rich manipulation tasks and three representative VLA backbones.
\end{itemize}












\section{Related Work}

\begin{figure*}[t]
    \centering
    \includegraphics[width=0.85\textwidth]{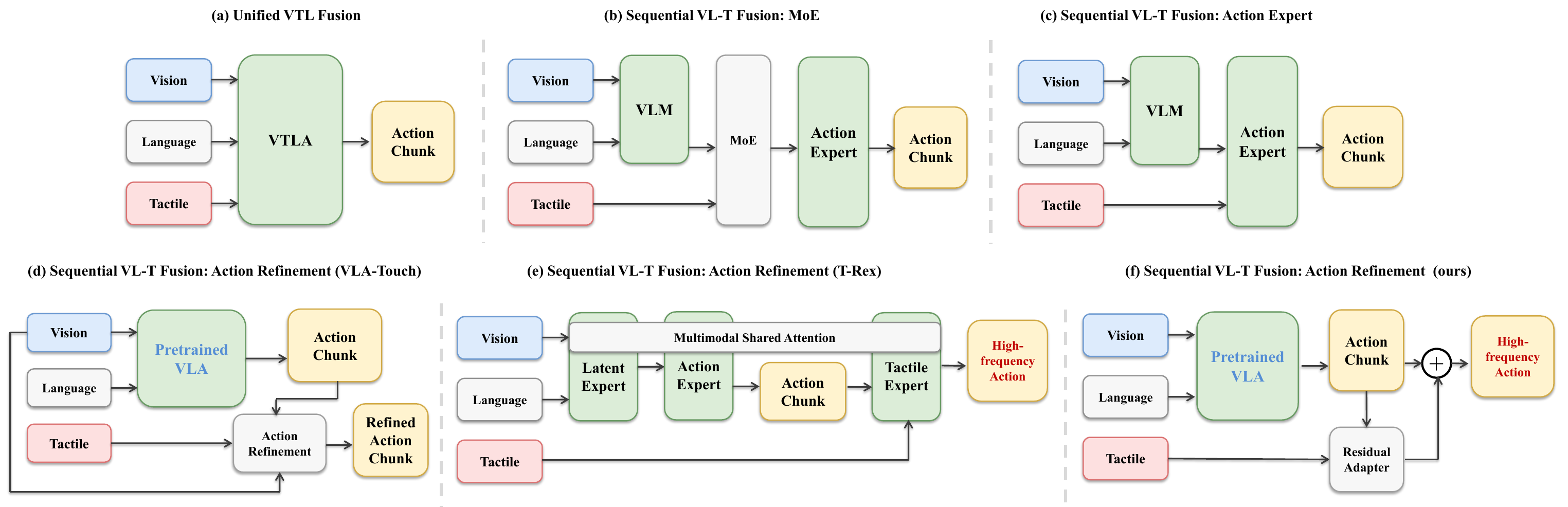}
    \caption{Representative multimodal fusion strategies in existing VTLA models. To our knowledge, only VT-Bridge (ours) retains pretrained VLA backbones without architectural modification while enabling high-frequency tactile adaptation.}
    \label{fig:sota}
\end{figure*}

\subsection{Vision-Language-Action Models.}
VLA models~\cite{brohan2022rt1,zitkovich2023rt2, driess2023palme, octo2024, kim2024openvla, kim2025finetuning} leverage large-scale vision-language pretraining to learn end-to-end robot policies that map visual observations and language instructions to low-level actions. Recent advances have evolved along several complementary architectural directions. Autoregressive models~\cite{kim2024openvla, kim2025finetuning,pertsch2025fast} formulate robot control as next-token prediction over discretized action tokens, naturally inheriting the training paradigm of vision-language models. Diffusion-based~\cite{chi2023diffusionpolicy, guo2024prediction,octo2024,liu2025hybridvla, zhou2025chatvla} approaches instead model continuous action distributions, producing smoother and more expressive behaviors at the cost of iterative inference. Flow-based methods, exemplified by $\pi_0$~\cite{black2024pi0}, replace diffusion sampling with conditional flow matching to significantly improve inference efficiency, while dual-system architectures such as $\pi_{0.5}$~\cite{black2025pi05} and GR00T-N1~\cite{nvidia2025gr00tn1} decouple high-level semantic reasoning from low-level action generation for long-horizon manipulation. Despite these advances, existing VLA models primarily emphasize semantic reasoning and visual understanding. Their robustness in contact-rich manipulation remains limited, motivating the development of tactile-aware VLA models.

\subsection{Vision-Tactile-Language-Action Models}

To remedy this gap, VTLA models have emerged as a promising research direction. As illustrated in Fig.~\ref{fig:sota}, existing VTLA methods can be broadly categorized according to how tactile information is fused:
\begin{itemize}
    \item \textit{Unified VTL Fusion:} As the most intuitive approach (Fig.~\ref{fig:sota}(a)), methods in this category independently encode vision, touch, and language  and feed the resulting representations into a unified multimodal transformer~\cite{cheng2025omnivtla, huang2025tactilevla, zhang2025vtla, gubernatorov2026hapticvla, zhang2026tacvla, zhao2026fdvla}. 
    \item \textit{Sequential VL-T Fusion:} Visual and language inputs are first jointly encoded, after which tactile information is incorporated at different stages. Tactile features can be fused through a Mixture of Experts (MoE) before action prediction (Fig.~\ref{fig:sota}(b))~\cite{yu2025forcevla,li2026forcevla2}, injected into the action decoder or action expert during action generation (Fig.~\ref{fig:sota}(c))~\cite{li2026atvla,li2026favla,wang2026never,zhang2026craft,zhang2026feeling}, or used to refine previously predicted actions (Fig.~\ref{fig:sota}(d) and Fig.~\ref{fig:sota}(e))~\cite{bi2025vla, niu2026t}.
    \item \textit{Others:} Tactile information can be fused at multiple stages~\cite{zhang2026feeling}, converted into continuous soft prompts for VLA reasoning~\cite{wang2026tacmamba}, or used by a VLM to predict stiffness for compliant action execution~\cite{zhang2026compliantvla}.
\end{itemize}

Despite their promising performance on demonstrated contact-rich tasks, developing VTLA models remains costly due to the substantial vision-tactile data and computational resources required. VLA-Touch~\cite{bi2025vla} takes an initial step toward adapting a pretrained VLA into a VTLA (Fig.~\ref{fig:sota}(d)), but the broader potential of this adaptation paradigm remains largely underexplored.


\subsection{Slow--Fast Inference Scheme}
Visual and force sensing are distinct in sensing characteristics yet functionally complementary~\cite{siciliano2008springer}. Visual observations are high-dimensional, captured at a relatively low frequency, and informative of global scene context, whereas force observations are low-dimensional, available at a high frequency, and tightly coupled to local contact dynamics. To accommodate these modality-specific characteristics, a slow--fast inference scheme was introduced in RDP~\cite{xue2025reactive}, in which a slow policy predicts latent action chunks from vision, while a fast decoder uses the latest tactile feedback to generate executable actions. This scheme effectively leverages the respective strengths of both modalities. Nevertheless, slow--fast inference has seen limited adoption in VTLA models, with only a few studies, including T-Rex~\cite{niu2026t} (Fig.~\ref{fig:sota}(e)), FAVLA~\cite{li2026favla}, and UniTacVLA~\cite{zhang2026unitacvla}.









\section{Methods}
To address the aforementioned limitations, this work proposes VT-Bridge, a lightweight residual adapter that augments pretrained VLA models with tactile-conditioned action adaptation for contact-rich manipulation. VT-Bridge retains the original pretrained VLA policy without modifying its structure and introduces a residual adapter working at a higher rate on top of its predicted action chunks, as illustrated in Fig.~\ref{fig:system_overview}.

\subsection{Low-frequency Pretrained VLA Backbone } 
The pretrained VLA backbone provides general-purpose visual-language understanding and guides the robot to complete the target manipulation task~\cite{black2024pi0, black2025pi05}. Given the current observation $\bm{o}$ and language instruction $\bm{l}$, the VLA policy $\pi^\mathrm{VLA}$ predicts an action chunk:
\begin{equation}
\bm{A} \sim \pi^\mathrm{VLA}(\cdot \mid \bm{o}, \bm{l}), 
    \label{eq:vla}
\end{equation}
where $\bm{A} = [\bm{a}_1, \bm{a}_{2}, \dots, \bm{a}_{H}]$ contains $H$ future actions. This predicted chunk captures  the coarse, task-level motion and provides the nominal action sequence subsequently refined by the residual adapter. 


\subsection{High-Frequency Residual Adapter}
The residual adapter refines the nominal motion predicted by the VLA backbone using tactile feedback. As depicted in Fig.~\ref{fig:system_overview}, we adopt a lightweight architectural design to facilitate faster inference and online adaptation to varying contact conditions. 

Specifically, the residual adapter consists of two parallel transformer encoders that separately process the VLA-predicted action chunk and latest force observations:  (i) The action encoder processes the embedded action chunk and produces an action feature $\bm{z}^a$ that represents the VLA's coarse motion intention and the overall action trend within the chunk. This feature is computed once and reused throughout chunk execution. (ii) In parallel, the force encoder extracts temporal contact features $\bm{z}^f_t$ from the latest force window:
\begin{equation}
    \bm{F}_{t} = [\bm{f}_{t-n+1},\bm{f}_{t-n+2}, \dots , \bm{f}_{t}],
\end{equation}
where $n$ denotes the window size, and $t$ indicates the current time step.  The encoded action feature $\bm{z}^a$ and force feature $\bm{z}_t^f$ are then concatenated and passed to a multilayer perceptron decoder to predict the residual action:
\begin{equation}
\bm{a}^{\mathrm{res}}_{t}
={\mathrm{MLP}}
\left(
[\bm{z}^a;\bm{z}_t^f]
\right).
\end{equation}
\subsection{Inference and Execution}
\vspace{-15pt}
\begin{figure} [h]
    \centering
    \includegraphics[width=1\linewidth]{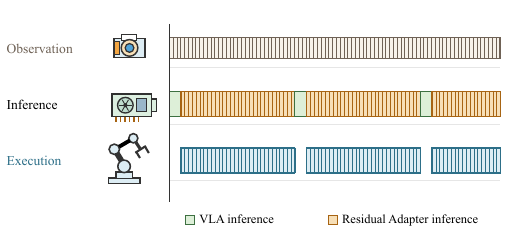}
    \caption{Temporal scheduling of the VLA backbone, residual adapter, and robot executor. 
    \label{fig:inferencescheduling}}
\end{figure}

During deployment, the VLA backbone and the residual adapter operate on the same GPU in a temporally interleaved manner, with each action chunk serving as the basic scheduling unit. As illustrated in Fig.~\ref{fig:inferencescheduling}, at the beginning of each cycle, the VLA backbone predicts an action chunk $\bm{A}$ via Eq.~\eqref{eq:vla}. The backbone then becomes idle, and VT-Bridge enters the adaptation-and-execution loop. At each execution step $t$, the residual adapter predicts an action adaptation from the complete action chunk and the latest force window: 
\begin{align}
    &\bm{a}^{\mathrm{res}}_{t} = \pi^\mathrm{res}(\cdot \mid \bm{A}, \bm{F}_{t}). 
\end{align}
The predicted residual action $\bm{a}^{\mathrm{res}}_{t}$ is then added to the corresponding nominal VLA action $\bm{a}_{t}$, yielding the adapted action $\bm{a}^{\mathrm{exe}}_{t}$ that is sent to the robot for execution. 
\begin{equation}
    \bm{a}^{\mathrm{exe}}_{t} = \bm{a}_{t}  + \bm{a}^{\mathrm{res}}_{t}.
\end{equation}

This loop continues until all actions in the predicted chunk have been executed. The residual adapter then pauses, and the VLA backbone is invoked to predict the next action chunk. In this manner, the two inference processes do not overlap and therefore do not contend for GPU computation.



\begin{figure*} [htbp]
    \centering
    \includegraphics[width=0.75\textwidth]{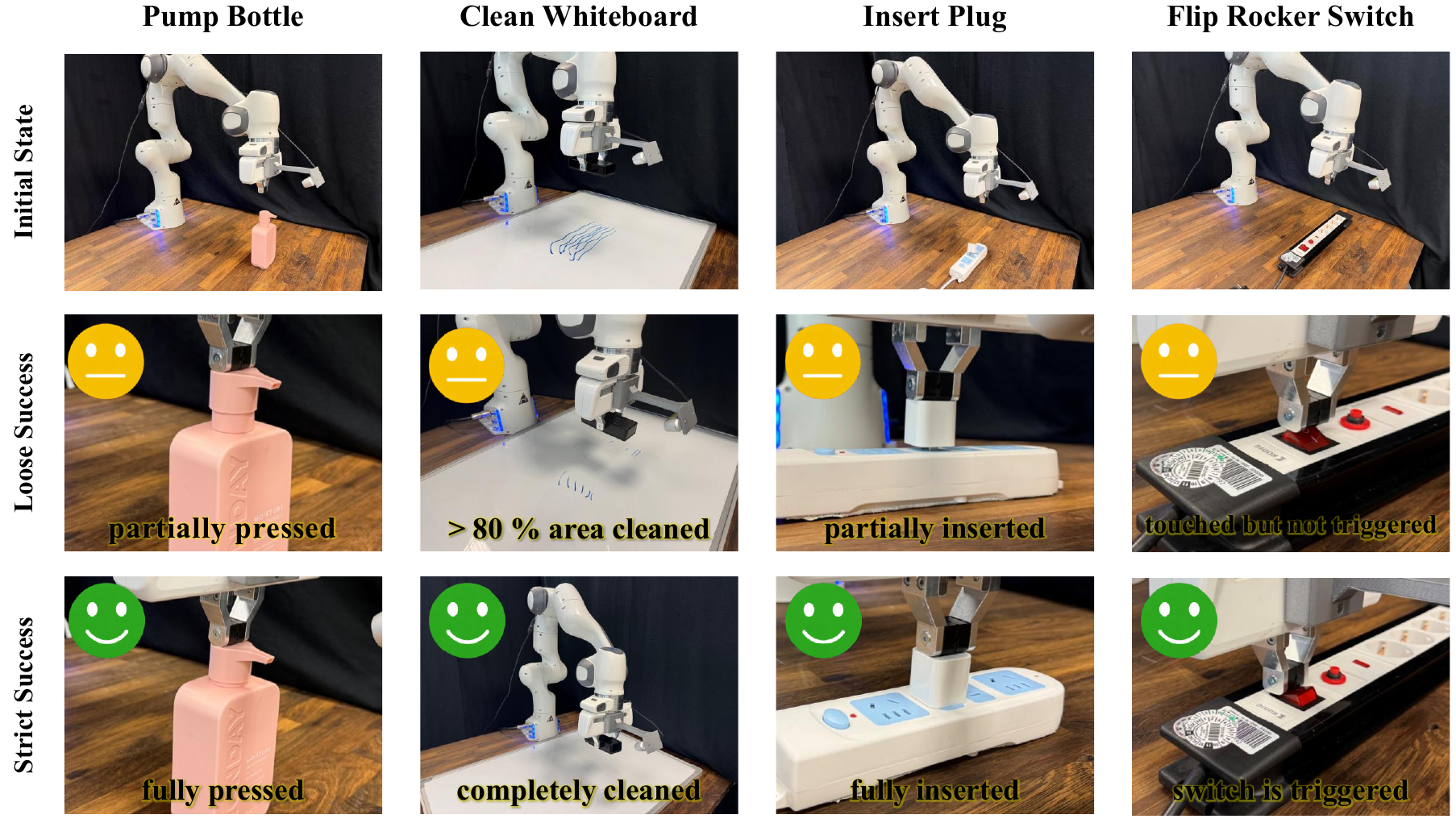}
    \caption{Overview of contact-rich manipulation tasks, showing representative initial states and examples satisfying the loose and strict success criteria.}
    \label{fig:task_config}
\end{figure*}

\section{Training}
The training pipeline of our method consists of four steps.

\subsection{Teleoperated Demonstration Collection}
Initially, we collect demonstrations for contact-rich manipulation tasks using the bilateral teleoperation strategy introduced in SharedAssembly~\cite{wu2025sharedassembly}. 
By reflecting follower-side interaction loads to the leader robot, this strategy enables the operator to perceive contact changes and regulate the applied force throughout the demonstration. 
During each demonstration, we synchronously record the camera stream, the measured robot kinematic state and its corresponding desired command, and tactile signals in the form of the estimated external end-effector force and external joint torques.

\subsection{VLA Backbone Fine-Tuning}
Afterwards, we adapt the pretrained VLA backbone to the target manipulation tasks using the collected demonstrations.  The camera observations, language instructions and robot states are converted into the training format required by each backbone, following its original observation preprocessing and action representation. We then fine-tune the backbone with LoRA~\cite{hu2022lora}. The resulting task-adapted backbone produces the nominal action chunks used for residual dataset construction in the next step. 

\subsection{Residual Dataset Construction \label{sec:residual_dataset}}
Then, we run the fine-tuned VLA backbone on the recorded observations to generate nominal action chunks. After temporal alignment, the target residual action at each execution step $k$ is computed as: 
\begin{equation}
    \bm{a}^\mathrm{res}_k = \bm{a}^\mathrm{desired}_k - \bm{a}_k, 
\end{equation}
where $\bm{a}^\mathrm{desired}_k$ denotes the desired command of the teleoperated robot's controller, and $\bm{a}_k$ is the corresponding action predicted by the fine-tuned VLA backbone. Using the desired command rather than the executed action preserves the operator's intended motion under contact. The effect of this choice is evaluated in an ablation study in Sec.~\ref{sec:ablation}. 

Each residual target is paired with the corresponding nominal action chunk $\bm{A}$ and contact-feedback window $\bm{F}_k$, which serve as the inputs to the residual adapter. Following $\pi_0$~\cite{black2024pi0}, VT-Bridge supports both Cartesian- and joint-space representations. To ensure consistency between the action and tactile representations, Cartesian actions are paired with estimated external end-effector force, whereas joint actions use external joint torque as tactile feedback. 

Besides, we set the target residual action to zero for all non-contact samples. This encourages the adapter to preserve the VLA backbone's nominal decisions during free-space motion and focus its adaptation on regulating physical interactions after contact is established.

\subsection{Residual Adapter Training}
Finally, we freeze the fine-tuned VLA backbone and optimize only the residual adapter on the constructed residual dataset. This decoupled training preserves the nominal task behavior of the VLA backbone while enabling contact-conditioned residual adaptation.

\section{Experiments}
In this section, we conduct a comprehensive suite of real-world contact-rich manipulation experiments to address the following research questions:
\begin{itemize}
    \item \textbf{Q1:} Can a lightweight residual adapter improve pretrained foundation VLAs' performance in real-world contact-rich manipulation tasks? 
    \item \textbf{Q2:} How does VT-Bridge overcome the last-inch bottleneck between geometrical near-success and functional task completion? 
    \item \textbf{Q3:} How does using desired poses rather than executed poses to construct residual targets affect manipulation performance?
\end{itemize}

\subsection{Experimental Setup}
\begin{table}[h]
\centering
\caption{Configuration Details}
\label{tab:runtime}
\begin{tabular}{lc}
\toprule
\textbf{Setting} & \textbf{Value} \\
\midrule
Demonstration sampling rate       & $30\,\mathrm{Hz}$ \\
Execution frequency               & $30\,\mathrm{Hz}$ \\
Joint stiffness matrix            &  $\operatorname{diag}(300,300,300,200,100,60,30)$ \\
\midrule
Force-history window $n$          & $50$ \\
Action-chunk horizon $H$          & $50$ \\
VLA fine-tuning steps         & $30,000$ \\ 
VLA fine-tuning batch size    & $32$ \\
VLA fine-tuning time              & $\sim8\,\mathrm{h}$\\
\midrule
$\pi_0$ chunk inference           & $153.1\,\mathrm{ms}$ \\
$\pi_{0.5}$ chunk inference       & $163.8\,\mathrm{ms}$ \\
SmolVLA chunk inference           & $155.4\,\mathrm{ms}$ \\
Action-chunk encoding             & $0.4\,\mathrm{ms}$ \\
Residual action prediction & $0.7\,\mathrm{ms}$ \\
\bottomrule
\end{tabular}
\end{table}
\subsubsection{Robot Platform}
The experimental platform comprises a Franka Emika Panda manipulator and two Intel RealSense D435i cameras. The robot controller runs on a NUC with an Intel i7-10700 CPU, while the VLA model and the proposed residual policy are executed on a separate workstation equipped with an NVIDIA RTX 3090 GPU. The foundation VLA models are fine-tuned using four NVIDIA A100-SXM4 GPUs. The concrete configurations in training and inference are summarized in Table~\ref{tab:runtime}.



\subsubsection{Evaluation Tasks} 
As illustrated in Fig.~\ref{fig:task_config}, the proposed method is evaluated on four contact-rich manipulation tasks: Pump Bottle, Clean Whiteboard, Insert Plug, and Flip Rocker Switch. These tasks cover diverse forms of physical interaction, including pressing, sustained surface contact, high-precision insertion, and discrete force-triggered actuation.

\subsection{Evaluation Metrics \label{sec: metric}} 

Each task is evaluated using two success criteria. The \textbf{strict success criterion} corresponds to complete task execution, while the \textbf{loose success criterion} indicates that the policy reaches the task-relevant interaction state and establishes the intended contact, but does not necessarily complete the task. This distinction separates failures in reaching the intended contact from failures in the subsequent contact-rich manipulation stage, enabling a more fine-grained evaluation of policy performance. Table~\ref{tab:criteria} summarizes the corresponding definitions.

\vspace{5pt}
\begin{table}[h]
    \centering
    \caption{Success Criteria \label{tab:criteria}}
    \begin{tabular}{lll}
         \toprule
         \textbf{Task}&  \textbf{Strict Criterion}& \textbf{Loose Criterion}\\
         \midrule
         Pump Bottle & Full press & Press only\\
         Clean Whiteboard& Complete cleaning &  $>80$\% area cleaned\\
         Insert Plug & Full insertion & $>80$\% insertion depth\\
         Flip Rocker Switch & Switch triggered & Contact only \\
         \bottomrule
    \end{tabular}
    
    \label{tab:placeholder}
\end{table}

For each task, we report the strict success rate  $\mathrm{SR}_{\mathrm{strict}}$, the loose success rate
$\mathrm{SR}_{\mathrm{loose}}$, and the conditional completion rate $\rho_\mathrm{sl}$:
\begin{equation}
\begin{aligned}
    \mathrm{SR}_\mathrm{strict} = N_{\mathrm{strict}}/N &\; (\%),\\
    \mathrm{SR}_\mathrm{loose} = N_{\mathrm{loose}}/N &\; (\%),\\
    \rho_{\mathrm{sl}}= N_\mathrm{strict}/N_{\mathrm{loose}} &\; (\%),
\end{aligned}
\end{equation}
where $N$ denotes the total number of trials, while $N_{\mathrm{strict}}$ and $N_{\mathrm{loose}}$ denote the numbers of
trials satisfying the strict and loose success criteria, respectively. $\rho_\mathrm{sl}$ measures the probability of achieving complete task execution
once the policy has already reached a near-success state. It therefore provides a focused measure of the policy's ability to complete the
contact-intensive stage of the task.

\subsection{Compared Methods}
We evaluate our approach on three representative foundation VLA backbones, namely $\pi_0$~\cite{black2024pi0}, $\pi_{0.5}$~\cite{black2025pi05}, and SmolVLA~\cite{shukor2025smolvla}. For each backbone, we compare three policy variants to isolate the effect of tactile integration. 
\begin{enumerate}
    \item \textbf{Fine-tuned}: the original VLA model fine-tuned with LoRA~\cite{hu2022lora}.

    \item \textbf{VLA-Touch}~\cite{bi2025vla}: we adopt the \emph{VLA Action Refinement with Touch} module proposed in VLA-Touch to augment the fine-tuned VLA backbone.

    \item \textbf{VT-Bridge} (ours): we employ the proposed lightweight residual adapter to augment the fine-tuned VLA backbone with the ability to regulate contact dynamics.
\end{enumerate}
For fair comparison, all methods use the same set of demonstrations and identical training settings unless otherwise specified.

\subsection{Experimental Procedure}

For each manipulation task, we collect 50 human demonstrations via bilateral teleoperation~\cite{wu2025sharedassembly} to construct the target manipulation dataset. For each pretrained VLA backbone, the model is first fine-tuned independently using LoRA on the collected demonstrations, serving as the Fine-tuned baseline. Based on the corresponding fine-tuned backbone, VLA-Touch\cite{bi2025vla} and our VT-Bridge are trained independently using the same demonstrations. The resulting backbone--method combinations are then evaluated on all four manipulation tasks. For each task, every combination is evaluated in 20 independent trials. Each trial is terminated when the strict success criterion is satisfied or the 90\,s time limit is reached.

\begin{table*}[]
\centering
\caption{Strict Success Rate~(\%) \label{tab:sc_strict}}
\begin{tabular}{c|c|ccccc}
\toprule
\textbf{Backbone} & \textbf{Method}       & \textbf{Pump Bottle} &\textbf{ Clean Whiteboard} &\textbf{ Insert Plug }& \textbf{Flip Rocker Switch} & \textbf{Average} \\
\midrule
\multirow{3}{*}{$\pi_0$~\cite{black2024pi0}}         & Fine-tuned~\cite{hu2022lora}                    &     0.0 (0/20)&     5.0 (1/20)&       0.0 (0/20)&    20.0 (4/20)& 6.3\\
                             & VLA-Touch~\cite{bi2025vla}           &   0.0 (0/20)&     5.0 (1/20)&        0.0 (0/20)&    50.0 (10/20)&     13.8 \\
                             & VT-Bridge (ours)   &  \textbf{85.0 (17/20)}&      \textbf{20.0 (4/20)}&      \textbf{70.0 (14/20)}&     \textbf{85.0 (17/20)}&   \textbf{65.0 }\\
\midrule
\multirow{3}{*}{$\pi_{0.5}$~\cite{black2025pi05}}         & Fine-tuned~\cite{hu2022lora}                     &  0.0 (0/20)& 35.0 (7/20)&   0.0 (0/20)&  55.0 (11/20)& 22.5 \\
                             & VLA-Touch~\cite{bi2025vla}          &0.0 (0/20)&         20.0 (4/20)&      5.0 (1/20)&          55.0 (11/20)&         20.0 \\
                             & VT-Bridge (ours)   &   \textbf{90.0 (18/20)}&  \textbf{40.0 (8/20)}&    \textbf{70.0 (14/20)}&   \textbf{75.0 (15/20)}&         \textbf{68.8}\\
\midrule
\multirow{3}{*}{SmolVLA~\cite{shukor2025smolvla}}         & Fine-tuned~\cite{hu2022lora}                     &  0.0 (0/20)          & 10.0 (2/20)           &   0.0 (0/20)          &  15.0 (3/20)        & 6.3       \\
                             & VLA-Touch~\cite{bi2025vla}           &   0.0 (0/20)&           5.0 (1/20)&  0.0 (0/20)&  25.0 (5/20)         &      7.5   \\
                             & VT-Bridge (ours)   &  \textbf{70.0 (14/20)}& \textbf{45.0 (9/20)}&  \textbf{65.0 (13/20)}& \textbf{40.0 (8/20)}           &         \textbf{55.0}\\
\bottomrule
\end{tabular}
\vspace{8pt}
\end{table*}

\begin{table*}[]
\centering
\vspace{-1pt}
\caption{Conditional Completion Rate $\rho_\mathrm{sl}$~(\%) \label{tab: rho} }
\begin{tabular}{c|c|ccccc}
\toprule
\textbf{Backbone} & \textbf{Method}       & \textbf{Pump Bottle} &\textbf{ Clean Whiteboard} &\textbf{ Insert Plug }& \textbf{Flip Rocker Switch} & \textbf{Average} \\
\midrule
\multirow{3}{*}{$\pi_0$~\cite{black2024pi0}}         & Fine-tuned~\cite{hu2022lora}                    &     0.0 (0/19)&     5.3 (1/19)&       0.0 (0/13)&    23.5 (4/17)& 7.2\\
                             & VLA-Touch~\cite{bi2025vla}           &   0.0 (0/5)&     5.9 (1/17) &        0.0 (0/16) &    66.7 (10/15) &     18.1\\
                             & VT-Bridge (ours)   &  \textbf{89.5 (17/19)}&     \textbf{ 20.0 (4/20)}&      \textbf{93.3 (14/15)}&     \textbf{100.0 (17/17)}&   \textbf{75.7}\\
\midrule
\multirow{3}{*}{$\pi_{0.5}$~\cite{black2025pi05}}         & Fine-tuned~\cite{hu2022lora}                     &  0.0 (0/20)& 35.0 (7/20)&   0.0 (0/16)&  \textbf{100.0 (11/11)}& 33.8\\
                             & VLA-Touch~\cite{bi2025vla}          &0.0 (0/10)&         20.0 (4/20)&      6.3 (1/16)&          84.6 (11/13)&         27.7\\
                             & VT-Bridge (ours)   &   \textbf{94.7 (18/19)}&  \textbf{40.0 (8/20)}&    \textbf{77.8 (14/18)}&    \textbf{100.0 (15/15)}&         \textbf{78.1}\\
\midrule
\multirow{3}{*}{SmolVLA~\cite{shukor2025smolvla}}         & Fine-tuned~\cite{hu2022lora}                     &  0.0 (0/19)& 10.5 (2/19)&   0.0 (0/8)          &  16.7 (3/18)       & 6.8       \\
                             & VLA-Touch~\cite{bi2025vla}           &   0.0 (0/4)&           6.3 (1/16)&  0.0 (0/10)&  29.4 (5/17)        &  8.9      \\
                             & VT-Bridge (ours)   &  \textbf{87.5 (14/16)}& \textbf{45.0 (9/20)}&  \textbf{86.7 (13/15)}& \textbf{61.5 (8/13)}           &      \textbf{70.2}   \\
\bottomrule
\end{tabular}
\end{table*}

\section{Experimental Results}
\subsection{Overall Performance on Contact-Rich Manipulation (Q1)}
\begin{figure} [b]
    \centering
    \vspace{-10pt}
    \includegraphics[width=1\linewidth]{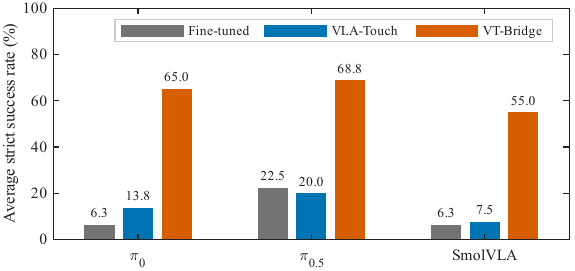}
    \caption{Average strict success rates across VLA backbones and real-world contact-rich manipulation tasks.}
    \label{fig:average_success_rate}
\end{figure}
As illustrated in Fig.~\ref{fig:average_success_rate}, VT-Bridge substantially outperforms baseline methods across all three representative VLA backbones. Compared with directly fine-tuning the backbones on the same robot demonstrations, VT-Bridge increases the overall average strict success rate from 11.7\% to 62.9\%, yielding an absolute improvement of 51.2 percentage points. The consistently substantial gains across pretrained VLA models demonstrate the broad effectiveness and applicability of VT-Bridge across various pretrained VLA backbones.

\begin{figure*}[h]
    \centering
    \includegraphics[width=0.9\textwidth]{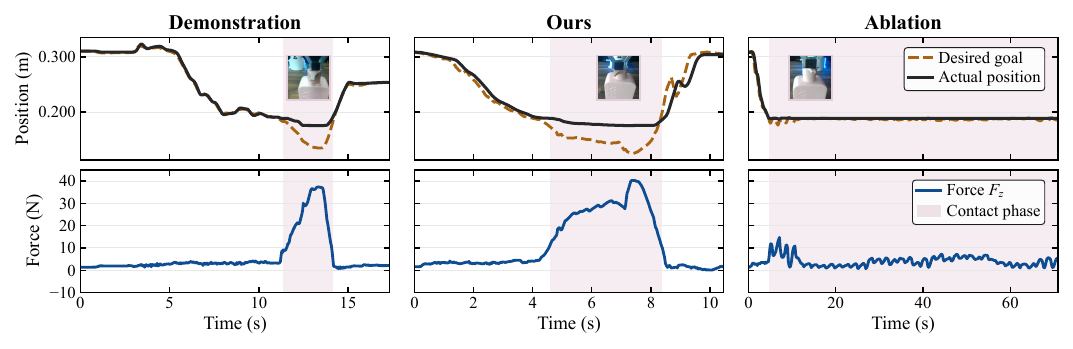}
    \vspace{-6pt}
    \caption{Interaction profiles along the \(z\)-axis for the Pump Bottle task.}
    \label{fig:ablation_bottle}
\end{figure*}

To further examine the performance across different contact-rich manipulation scenarios, Table~\ref{tab:sc_strict} reports the strict success rate for each individual task. Notably, VT-Bridge achieves the highest strict success rate in all task–backbone combinations, without exception. This remarkably consistent advantage highlights the robustness and broad applicability of the proposed residual adaptation framework. 

\subsection{Last-Inch Bottleneck in Contact-Rich Manipulation (Q2)}
Our experiments reveal that a notable proportion of failure cases do not result from an incorrect task intention. Instead, the policies often approach task completion but fail at the final contact-sensitive step. For example, the robot may successfully depress the pump head but not far enough to dispense liquid, or partially insert the plug without fully seating it. Although these executions are geometrically close to success, they do not achieve the intended functional outcome and must therefore be considered failures in real-world manipulation. We refer to this gap between geometrical near-success and functional completion as the \textbf{last-inch bottleneck in contact-rich manipulation}. 

To quantify a policy’s ability to overcome this bottleneck, we analyze the conditional completion rate \(\rho_{\mathrm{sl}}\), introduced in Sec.~\ref{sec: metric}. By conditioning on loose success, this metric reduces the influence of backbone-dependent differences in task understanding and motion generation, enabling a more focused evaluation of tactile adaptation during last-inch completion.

As depicted in Table~\ref{tab: rho}, VT-Bridge achieves the highest conditional completion rate in all task–backbone combinations. On average, VT-Bridge increases the macro-averaged conditional completion rate by 58.8 percentage points over fine-tuned VLA models, demonstrating a substantially stronger ability to overcome the last-inch bottleneck. The Pump Bottle task provides a striking example: none of the baseline–backbone combinations successfully converted a geometrical near-success into functional completion, as they failed to fully depress the pump head. In contrast, VT-Bridge achieves an average conditional completion rate of 90.6\%. A similar trend is observed in the Insert Plug task, where fine-tuned VLAs remain at 0\% across all three backbones, and VLA-Touch reaches only 6.3\% in one combination, while VT-Bridge achieves an average of 85.9\%. These results show that our method specifically improves the final transition from geometric near-success to functional completion.

\subsection{Ablation Study on Residual Target Construction: Desired vs. Executed Poses (Q3)~\label{sec:ablation}  }
\begin{figure}[b]
\vspace{-9pt}
    \centering
    \includegraphics[width=1\linewidth]{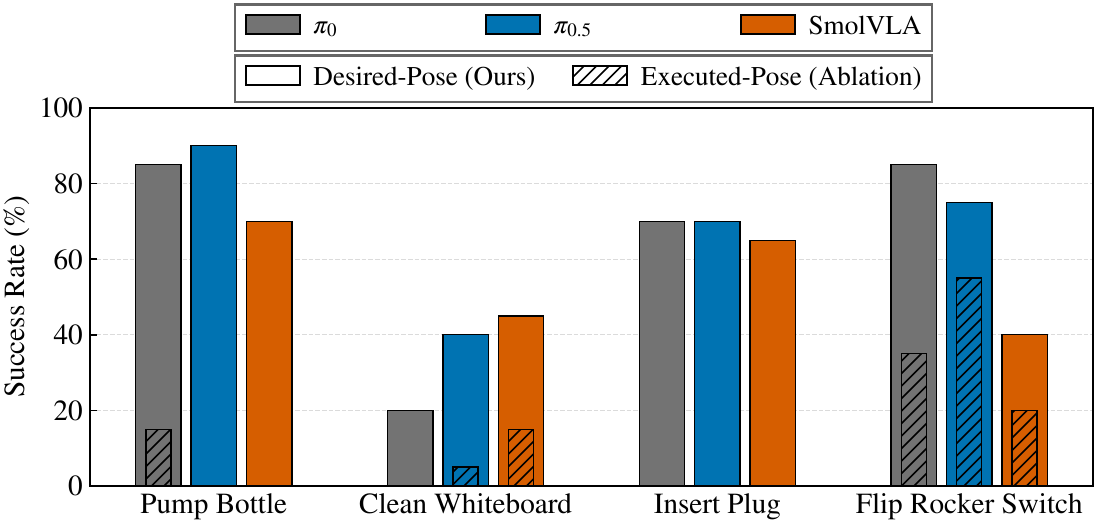}
    \caption{Ablation of residual target construction. Strict success rates obtained using desired-pose (ours) and executed-pose (ablation) to construct the residual dataset.}
    \label{fig:ablation_bar_chart}
\end{figure}

As introduced in Sec.~\ref{sec:residual_dataset}, our default residual targets are computed as the difference between the desired poses $\bm{a}^\mathrm{desired}$ recorded in the demonstrations and the nominal poses $\bm{a}$ predicted by the fine-tuned VLA. For this ablation, we construct an alternative residual dataset by replacing the desired poses with the corresponding executed poses $\bm{a}^\mathrm{executed}$, while keeping all other settings unchanged. As shown in Fig.~\ref{fig:ablation_bar_chart}, residual targets constructed from desired poses consistently outperform those constructed from executed poses across all task–backbone combinations. On average, using desired poses increases the strict success rate from \(12.1\%\) to \(62.9\%\).

To support safe physical interaction, we use impedance control throughout our experiments. In tasks requiring high interaction forces, controller compliance creates a pronounced offset between the desired and executed poses. As illustrated in Fig.~\ref{fig:ablation_bottle}, fully depressing the pump head requires a peak interaction force of approximately $40~\mathrm{N}$ in both the demonstration and the successful VT-Bridge trial. Correspondingly, the desired–executed pose offset reaches approximately $4\text{--}5~\mathrm{cm}$. Residual targets constructed from desired poses explicitly incorporate this force-related offset into the training data, whereas those constructed from executed poses omit it. Consequently, even if the ablation policy accurately predicts the executed pose recorded in the demonstration, issuing that pose as a new desired command produces another tracking shortfall under contact. The robot guided by the ablation policy therefore fails to reproduce the demonstrated pose and contact force. Although employing a high-stiffness controller could reduce this offset, it would also increase the risk of excessive interaction forces~\cite{hogan1985impedance,hogan2005impedance}. By contrast, our design improves task reliability while maintaining safe physical interaction.


A similar pattern is observed in Insert Plug, where the peak interaction force also approaches \(40~\mathrm{N}\). By contrast, Flip Rocker Switch requires only around \(7~\mathrm{N}\) to accomplish the task, resulting in a smaller compliance-induced offset. The executed-pose ablation therefore performs better on this task than on the two higher-force tasks. Nevertheless, constructing residual targets from desired poses still yields an 82\% relative improvement over the ablation policy on this task. Overall, using desired poses $\bm{a}^\mathrm{desired}$ to construct residual targets consistently improves performance across evaluated contact-rich tasks, with more pronounced gains as the required interaction force increases.

\section{Conclusion}

This work presents VT-Bridge, a resource-efficient residual adaptation strategy that bridges pretrained foundation VLAs to tactile-aware VTLAs for contact-rich manipulation. Rather than training a VTLA model from scratch or modifying the pretrained VLA backbone architectures, VT-Bridge employs an identical lightweight residual-adapter architecture across pretrained VLAs and uses backbone-specific weights to refine actions at the robot execution frequency. This design retains the vision-language understanding and motion-generation capabilities acquired through large-scale VLA pretraining while enabling responsive contact regulation. Using 50 vision-tactile demonstrations per task, VT-Bridge consistently improved three representative VLA backbones across four real-world contact-rich manipulation tasks, raising the average task completion rate from 11.7\% to 62.9\%. These experimental results demonstrate its effectiveness and broad applicability, showcasing a promising pathway for developing VTLAs from existing pretrained VLAs. Future work will focus on improving the representation and adaptation capabilities of the residual adapter to achieve stronger and more consistent performance gains.

\section*{ACKNOWLEDGMENT}
As non-native English speakers, the authors used GPT-5.6 for grammatical and linguistic refinement. Codex was also used to organize the codebase for open-source release. All AI-assisted suggestions and modifications were reviewed and verified by the authors.

\bibliography{IEEEabrv,mybib2022}
\bibliographystyle{myIEEEtran}


\label{last-page}
\end{document}